\documentclass[10pt,twocolumn,letterpaper]{article}

\usepackage{cvpr}
\usepackage{times}
\usepackage{epsfig}
\usepackage{graphicx}
\usepackage{amsmath}
\usepackage{enumerate}
\usepackage{amssymb}
\usepackage{enumitem}
\usepackage{xcolor}
\usepackage{booktabs} 
\usepackage{gensymb} 
\usepackage{lipsum} 
\usepackage{graphbox} 
\usepackage{blindtext}
\usepackage{nameref}
\usepackage{pythonhighlight}

\usepackage[pagebackref=true,breaklinks=true,letterpaper=true,colorlinks,bookmarks=false]{hyperref}

\cvprfinalcopy 

\begin{document}

\title{Visual Physics: Discovering Physical Laws from Videos}

\author{
{
Pradyumna Chari$^\dag$,
Chinmay Talegaonkar$^\dag$,
Yunhao Ba$^\dag$ \& 
Achuta Kadambi
}\\
Electrical and Computer Engineering Department\\
University of California, Los Angeles (UCLA)\\
{\tt\small \{pradyumnac, chinmay0301, yhba, achuta\}@ucla.edu}
}

\maketitle

\renewcommand*{\thefootnote}{$\dag$}
\setcounter{footnote}{1}
\footnotetext{Indicates equal contribution.}
\renewcommand*{\thefootnote}{\arabic{footnote}}
\setcounter{footnote}{0}

\begin{abstract}
In this paper, we teach a machine to discover the laws of physics from video streams. We assume no prior knowledge of physics, beyond a temporal stream of bounding boxes. The problem is very difficult because a machine must learn not only a governing equation (e.g. projectile motion) but also the existence of governing parameters (e.g. velocities). We evaluate our ability to discover physical laws on videos of elementary physical phenomena, such as projectile motion or circular motion. These elementary tasks have textbook governing equations and enable ground truth verification of our approach.
\end{abstract}
 \section{Introduction}

\begin{figure}
    \centering
    \includegraphics[width=\linewidth]{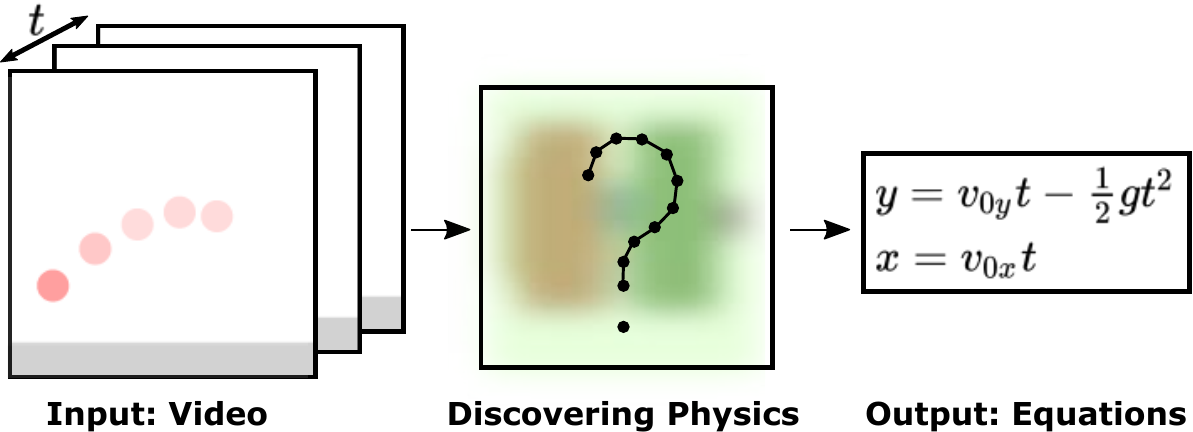}
    \caption{
    \textbf{Discovering physical equations from visual cues without human intervention.} Here, we show- how an input video of projectile motion can be processed by our method to recover both the governing equation of motion, as well as two governing parameters of initial velocities (both horizontal and vertical).}
    \label{fig:general_idea}
\end{figure}

This paper aims to teach a machine to discover the laws of physics from video streams. In the apocryphal story, Isaac Newton's observation of a falling apple was a catalyst for deriving his physical laws. In like fashion, our machine aims to observe the dynamics of a moving object as a means to infer physical laws. We refer to this as \emph{discovering physics from video}, as shown in Figure~\ref{fig:general_idea}.

The discovery problem is very difficult because a machine must derive not only the governing equations of a physical model but also governing parameters like velocity. We emphasize that a discovery algorithm like ours does not know \emph{a priori} what ``velocity'' means---it must learn the existence of velocity. In order to handle the underdetermined nature of recovering both governing equations and governing parameters, we make a few assumptions. Section~\ref{sec:discovery_definition} expands on our assumptions, which we believe are the most relaxed to date. 

Our work is powered by methods from representation learning and evolutionary algorithms. The discovery of underlying governing parameters is achieved using a modified $\beta$-variational autoencoder ($\beta$-VAE) to obtain latent representations. These are then used in an equation discovery step, driven by genetic programming approaches. Our approach is able to learn equations that symbolically match ground truth, and have governing parameters that correspond to human interpretable constructs (e.g. velocity, angular frequency).

\paragraph{Contributions:} Our key contribution is a first attempt at an algorithm that is able to re-discover both governing equations and governing parameters from video. Previous work can either discover governing equations or the parameters, but not both. We test the algorithm on both synthetic data (with and without noise), as well as real data. Our performance analysis shows that the proposed method results in symbolically accurate expressions, and interpretable governing parameter discovery for a variety of simple, yet fundamental physics tasks. The method is also found to be robust to large amounts of positional noise and effective under a range of input data sizes. To lay a foundation for future work, we release the Visual Physics dataset, consisting of both real and synthetic videos of dynamic physical phenomena.

\section{Related Work}
\label{sec:related}

Although our goals are different, we are inspired by work in physics-based computer vision, physical representation learning, and symbolic equation derivation. 

\paragraph{Physics-based computer vision} encompasses the use of known physical models to either directly solve or inspire computer vision techniques. Techniques like shape from shading~\cite{ikeuchi1981numerical,horn1989shape} and photometric stereo~\cite{woodham1980photometric} use known models of optical physics to estimate shape. Along this theme, recent work in the area of computational light transport has advanced the field to see around corners~\cite{ramesh20085d,velten2012recovering,o2018confocal,xin2019theory} or infer material properties~\cite{tanaka2018material}.\footnote{For an overview of the physics of light transport, the reader is directed to an ACM SIGGRAPH course by O'Toole and Wetzstein~\cite{o2014computational}.} Known physical models can also be used to inspire the design of vision algorithms. Examples include deformable parts models~\cite{felzenszwalb2008discriminatively,felzenszwalb2009object} or snakes~\cite{kass1988snakes}, which use the physics of springs to design computer vision cost functions. The recent popularity of data-driven techniques has spawned a family of work that combines a known physical model with pattern recognition. For example,~\cite{gregor2010learning, diamond2017unrolled} unfold the existing physical models as the backbone in the network architecture; \cite{chen2018reblur2deblur, stewart2017label} use physical information to supervise the training process; \cite{fei2019geo} relies on gravity cues to improve depth estimation; and  \cite{davis2015visual, jin2017deep, kang2017deep, ba2019physics, li2019restoration, Halder_2019_ICCV, zeng2019tossingbot} introduce physics-based learning to set the new state-of-the-art in a range of vision problem domains. These approaches are powered by knowledge of a physical model, whereas our work has the complementary aim of learning the underlying model. 

\paragraph{Learning physical parameters from visual inputs} has been a topic of interest in recent years. For instance,~\cite{JiajunWu2015Gallileo, Brubaker2009, Bhat2002, Mottaghi15Newton, purushwalkam2019bounce, Wu2017Deanimation} estimate parameters or equivalent information for well-characterized physical equations with visual inputs. These can be incorporated into realistic physical engines to infer complex system behavior. Fragidaki et al.~\cite{Fragidaki16Billiards} integrate the model of external dynamics within the agent to play simulated billiards games. More recently,~\cite{Battalgia2016IN,Watters2017VisualInteractionNetworks} deploy interaction networks with graph inputs to encode the interactions among objects in complex environments, and estimate other invariant quantities of the phenomenon using deep learning. In the field of controls, Shi et al.~\cite{shi2019neural} learn the near-ground dynamics to achieve stable trajectory control. While these prior attempts are capable of predicting the system dynamics precisely, they also require a well-characterized physical model. 

\paragraph{Symbolic regression} aims to generate symbolic equations from a space of mathematical expressions to fit the distributions of input samples. Genetic programming~\cite{GeneticProgramming} is one of the prevalent methods in this field, with previous applications in discovering Lagrangians~\cite{hills2015algorithm} and nonlinear model structure identification~\cite{winkler2005new}. Additional features from the input variables~\cite{Kaizen, GP-RVM} and partial derivatives pairs~\cite{schmidt2009distilling} can also be introduced into genetic programming for more reliable regression. Other evolutionary methods can also be used to derive partial differential equations (PDEs)~\cite{maslyaev2019data}. Sparse regression~\cite{Brunton2016} and dimensional function synthesis~\cite{wang2019deriving} are two other alternatives to conduct symbolic regression. Recently, deep neural networks (DNNs) have also been utilized to generate symbolic regression~\cite{EQL, EQL_extented, NeuralSymbolicRegression2019}. These existing methods usually require predetermined terms or prior knowledge from physics. 

\begin{figure}
    \centering
    \includegraphics[width=\linewidth]{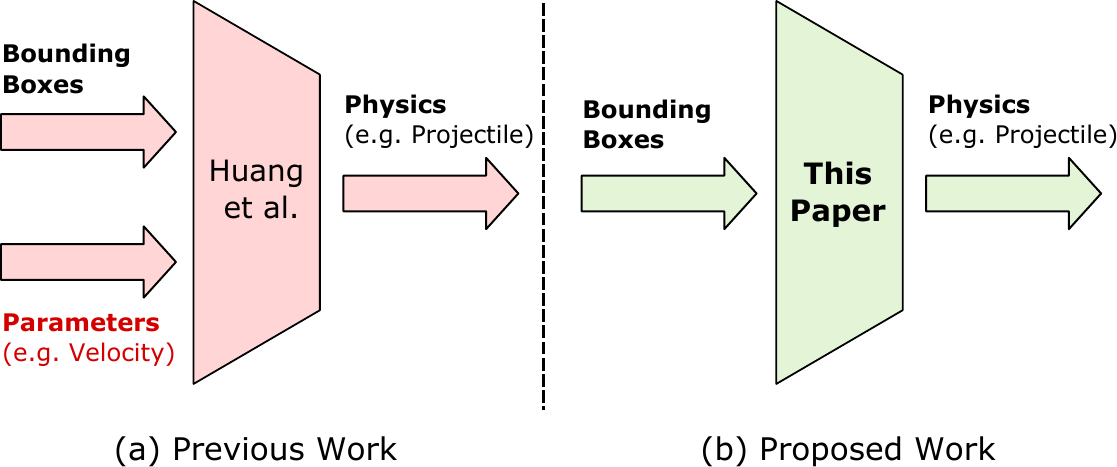}
    \caption{\textbf{Previous work~\cite{huang2018NIPSworkshop} (a) requires both a temporal stream of bounding boxes and the physical parameters.} (b) Our proposed technique also requires a stream of bounding boxes, but is able to discover latent parameters that correspond to true physical parameters, like velocity or angular frequency.}
    \label{fig:comparsion_with_others}
\end{figure}

\begin{figure*}
    \centering
    \includegraphics[width=\linewidth]{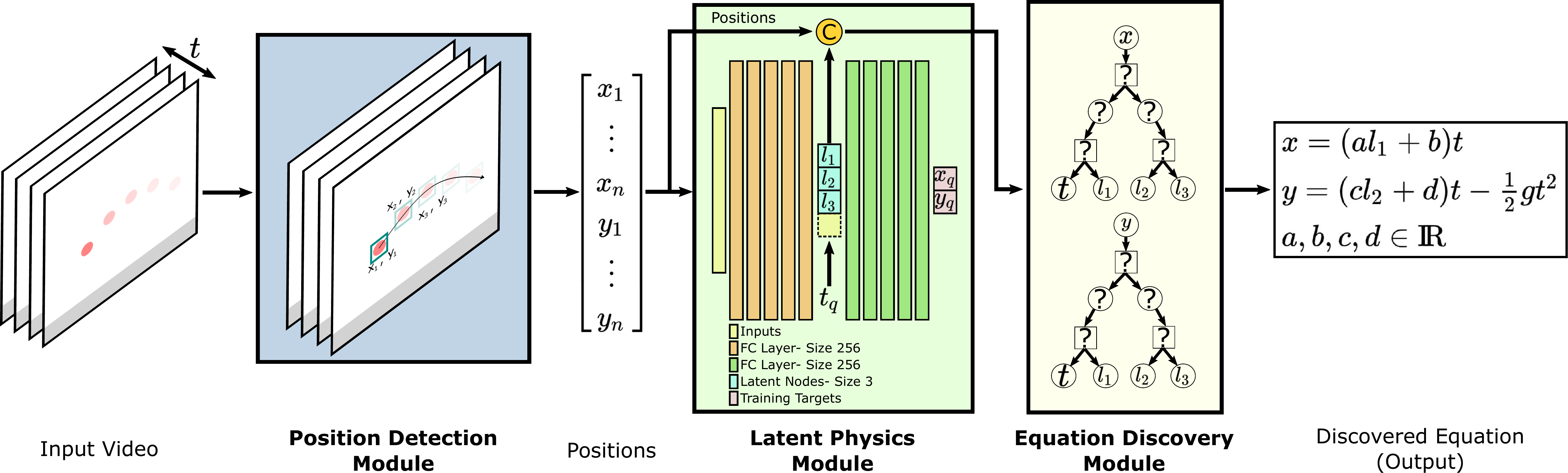}
    \caption{
    \textbf{An overview of the proposed Visual Physics framework.} We use a number of video clips as inputs to our system. The extracted position information is fed through the physics parameter extractor, which identifies the governing physical parameters for the phenomenon. These are used as inputs to the genetic programming step, in order to identify a human interpretable, closed form expression for the phenomenon. 
    }
    \label{fig:algorithm_pipeline}
\end{figure*}

\section{Defining Discovery and its Assumptions} \label{sec:discovery_definition}

\paragraph{Assumptions:} This paper represents only a first attempt to discover the laws of physics from video. As such, we make certain assumptions. First, we restrict our focus to the dynamics of single objects (rather than groups of objects). Second, it is assumed that we know the object for which we would like to derive the physical equations. Third, we assume that videos are in sequence. We believe these assumptions are sufficiently general to allow us to characterize our technique as ``discovering physics''. For example, the apocrypyhal story of Isaac Newton observing the apple falling aligns with the three assumptions outlined above. In the story, Newton was watching a temporal sequence of a single object in motion and was able to inductively reason about the laws of physics.

\paragraph{Defining ``discovery of physics'':} We define discovery of physics as discovering \emph{both} the governing parameters and governing equations. Given the assumptions from the previous paragraph, we must therefore discover all parameters except for the object location and time. As compared to Huang et al.~\cite{huang2018NIPSworkshop}, where the parameters of the governing equations are used as prior knowledge, our attempt at discovery is more general. Concretely, for a task like trajectory estimation, our framework has to tackle the challenging task of learning both the projectile equation, as well as the existence of a ``velocity'' term, from video input. Refer to Figure~\ref{fig:comparsion_with_others} for details. 

\section{Algorithm Architecture for Discovery}

Having defined ``discovery'' in Section~\ref{sec:discovery_definition}, we now describe a framework that enables discovery of physics from video. There are three interconnected modules that handle position detection, latent physics discovery, and equation discovery, respectively. Figure~\ref{fig:algorithm_pipeline} summarizes this framework.

\paragraph{Position detection module:} We build the Visual Physics framework based on the assumption that the underlying physical equations are reflected in the dynamics of an object across different time steps. Therefore, a robust object detection algorithm is required at the first stage to achieve accurate moving object localization for diversified categories of objects. We deploy a pretrained Mask R-CNN~\cite{he2017mask} to extract the bounding box of the object in each frame, and the centroid of the detected bounding box is considered as the object location in a particular frame.

\paragraph{Latent physics module:} The objective of the Visual Physics framework is to derive the governing physical laws without prior knowledge. To achieve this goal, we need to infer the associated latent governing parameters from positional observations. VAEs~\cite{kingma2013auto} have been widely deployed to extract the latent representations with applications in physics, such as SciNet~\cite{iten2018discovering}. We adopt a modified $\beta$-VAE architecture for our latent physics module as well. The encoder takes a vector corresponding to the object trajectory at uniformly sampled time instants as the input, and condenses them into a limited number of latent parameters. The decoder tries to reconstruct the object location $(x_q,y_q)$ at an unseen time instant with these latent parameters $[l_1$ $l_2$ $l_3]^T$ and the time instant $t_q$ as inputs. This module is supervised by the object locations without other prior physical knowledge. Once the network converges, both locations obtained from the position detection module, and the corresponding learned hidden representations from the latent physics module are paired as the equation discovery module input.

\paragraph{Equation discovery module:} We concatenate the latent parameters and positional observations, and use this as input to a symbolic regression approach. Vanilla genetic programming approaches are usually subject to convergence issues, and may lead to trivial equations that are not descriptive for the physics associated with the data. Schmidt et al.~\cite{schmidt2009distilling} alleviate this problem by introducing partial derivative pairs between the input variables as a search criterion. We follow this strategy to design an equation discovery module, capable of generating multiple equations with a range of equation complexity and fit accuracy. The final output is a symbolic equation that is Pareto-optimal.

\section{Implementation} \label{sec:impl}

\paragraph{Visual Physics dataset:} To evaluate the proposed framework, we generate both a real and synthetic dataset of videos covering physical phenomena. Table~\ref{tab:dataset} shows three simulated phenomena: \textsc{Free Fall}, \textsc{Constant Acceleration Motion} and \textsc{Uniform Circular Motion}. Each synthetic task includes 600 videos with randomly sampled physical parameters. We additionally include real video clips for \textsc{Free Fall} (411 videos). For all scenes, the physical phenomena is known in closed-form, enabling us to compare our proposed approach to ground truth. While the physics may seem elementary, we test in real-world conditions and add noise to make the task harder. Please see the supplement for additional scenes with a wider range of complexity. 


\begin{table*}
    \begin{center}
    \begin{tabular}{p{1.6cm} c p{12.5cm}}
    \toprule
    Dataset                                 & Visualization      & Description \\
    \midrule
    \textsc{free fall}                      & \raisebox{-1.3\totalheight}{\includegraphics[width=0.1\textwidth]{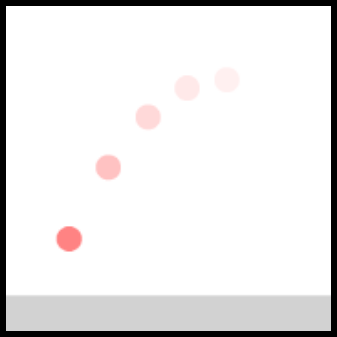}}
    
    & This dataset consists of 600 videos of 150 frames each at a frame rate of 240 frames per second. The frame size is chosen to be 720$\times$720 pixels. The object of interest is released with random initial velocities, from random points across different videos. The positions are selected from a uniform distribution, such that the initial position is in the bottom-left quadrant of the image. Initial velocities are also selected from a uniform distribution such that the object stays in the frame for the duration of the video. The object is acted upon by earth's gravity ($9.8m/s^2$ at a scale of 300 pixels per meter), which is the only active external agent.   \\
    \midrule
    
    \textsc{constant acceleration motion}                   & \raisebox{-0.93\totalheight}{\includegraphics[width=0.1\textwidth]{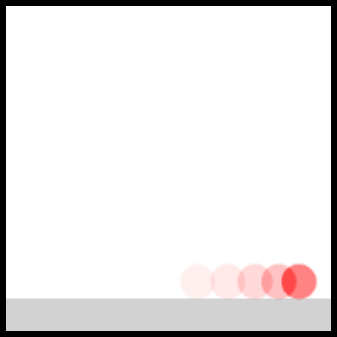}}
    
    & This dataset consists of 600 videos of 200 frames each, at a frame rate of 40 frames per second and a frame size of 720$\times$720 pixels. Here, the object of interest is released horizontally with a fixed initial velocity of $5 m/s$ (at a scale of 8 pixels per meter), and is acted upon by a uniformly random sampled external force, leading to an acceleration $a\in [0,4]$ $m/s^2$.  \\
    \midrule
    
    \textsc{uniform circular motion}                & \raisebox{-1.05\totalheight}{\includegraphics[width=0.1\textwidth]{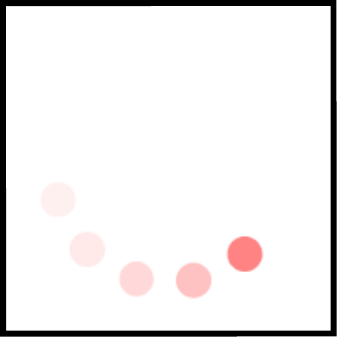}}
    
    & This dataset consists of 600 videos of 200 frames each, at a frame rate of 20 frames per second and a frame size of 720$\times$720 pixels. In this scenario, the object of interest is in uniform circular motion at a fixed radius of 5 m (at a scale of 50 pixels per meter), with angular velocity $\omega \in [1,2]$ rad/s. The center of rotation is kept fixed across all dataset videos. The initial position of the object is kept fixed, and no additional external force affects this motion (that is, the motion is assumed to be in the horizontal plane).\\
    \bottomrule
    \end{tabular}
    \end{center}
    \caption{\textbf{Description of the synthetic Visual Physics dataset.} These three physical phenomena are representative of fundamental trajectory motion. Although all scenes describe trajectories, the governing equations and parameters are different (e.g. polynomial for some, and sinusoidal for others).}
    \label{tab:dataset}
\end{table*}

\paragraph{Software implementation and training details:} For the position detection module, we deploy a Mask R-CNN~\cite{he2017mask} pretrained on COCO dataset~\cite{lin2014microsoft}. As to the physical inference module, both the encoder and the decoder consist of six fully-connected layers, and the size of the latent parameters is set to be three. We use the mean squared error (MSE) of the reconstructed locations and the $\beta$-VAE loss~\cite{higgins2017betaVAE} to supervise the training process. $\beta$-VAE penalty is introduced to encourage the disentanglement of latent representations, so that independent physical parameters are inferred in separate latent nodes. The entire loss function $L$ of the latent physics network can be written as follows:
\begin{equation}
    L = L_{mse}(Y_{t_q}, \hat{Y}_{t_q}) + \beta L_{kl}(Z),
\end{equation}
where $Y_{t_q}$ is the ground-truth location at time step $t_q$, $\hat{Y}_{t_q}$ is the estimated location from the network, $L_{mse}(\cdot)$ is the MSE loss, $Z$ denotes the extracted latent representations, $L_{kl}(\cdot)$ denotes the Kullback–Leibler divergence between a Gaussian prior, and $\beta$ is the balance factor for the $\beta$-VAE loss as described in~\cite{higgins2017betaVAE}. We use Adam optimizer~\cite{kingma2014adam} with an initial learning rate of 0.001, and this learning rate is decayed exponentially with a factor of 0.99 every 200 epochs. All the networks are implemented in the PyTorch framework~\cite{paszke2017automatic}. We construct the equation discovery module by using the widely available Eureqa package~\cite{EureqaSoftware}. The candidate operation set includes all the basic operations, such as addition, multiplication, and sine function. We search two equations for horizontal and vertical directions separately, and R-squared value is used to measure the goodness of fit during searching. Please refer to Appendix~\ref{app:software_implementation} for additional implementation details.

\section{Evaluation}

Section~\ref{ss:synth} evaluates our results on discovering equations from synthetic videos. Section~\ref{ss:real} shows that the method generalizes to real data. Finally, Section~\ref{ss:performance} tests the robustness of our technique by introducing noise and other confounding factors. 

\begin{figure*}
    \centering
    \includegraphics[width=\linewidth]{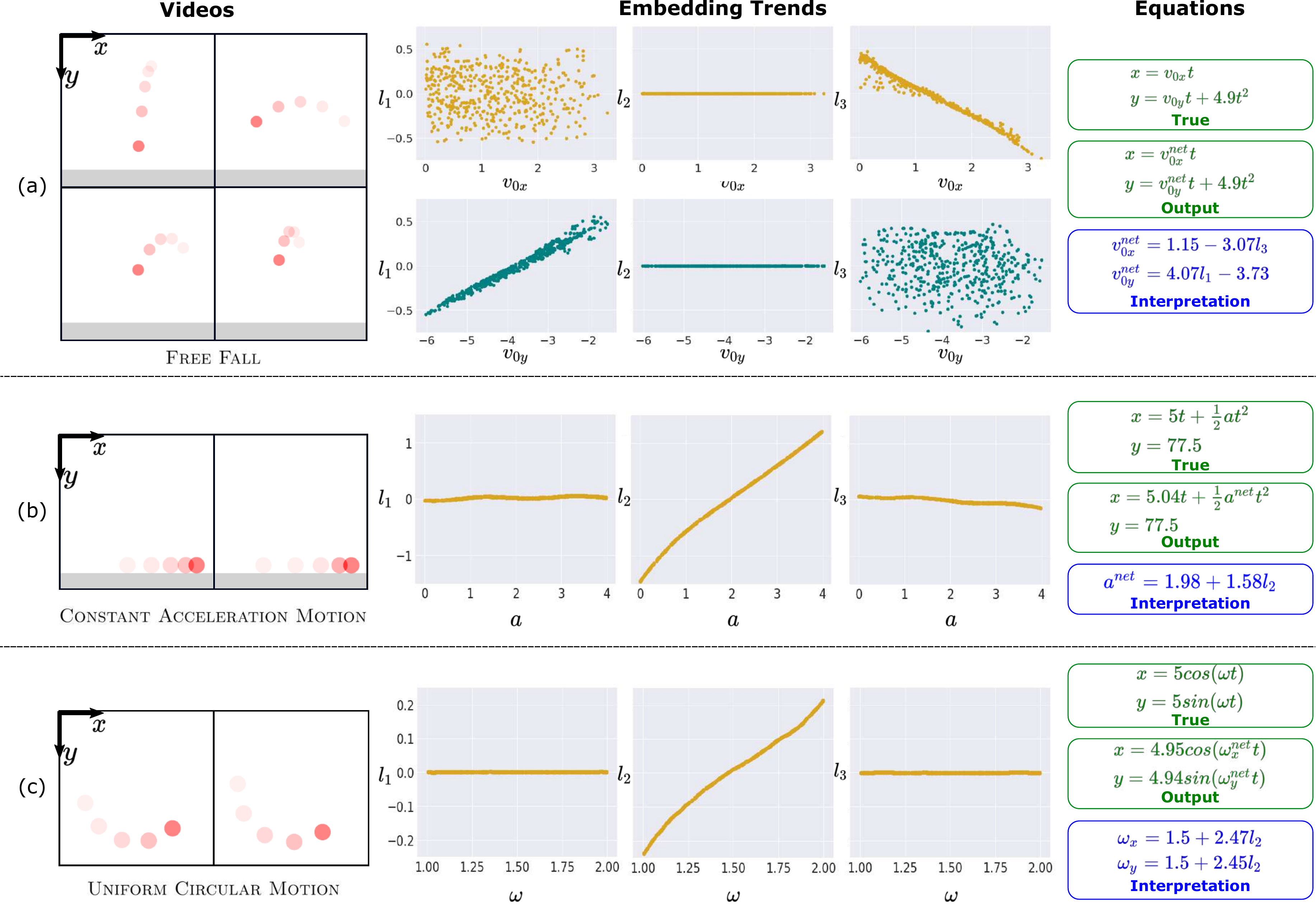}
    \caption{
    \textbf{Discovered physical equations from Visual Physics framework, on simulated videos.} We show the observed embedding trends and the obtained equations, which are both accurate in fitting to the observations as well as in human interpretable form. Results are shown on three simulated datasets: ball toss, acceleration and circular motion.
    }
    \label{fig:main_result}
\end{figure*}

\subsection{Synthetic Data Evaluation}
\label{ss:synth}

\begin{figure*}
    \centering
    \includegraphics[width=\linewidth]{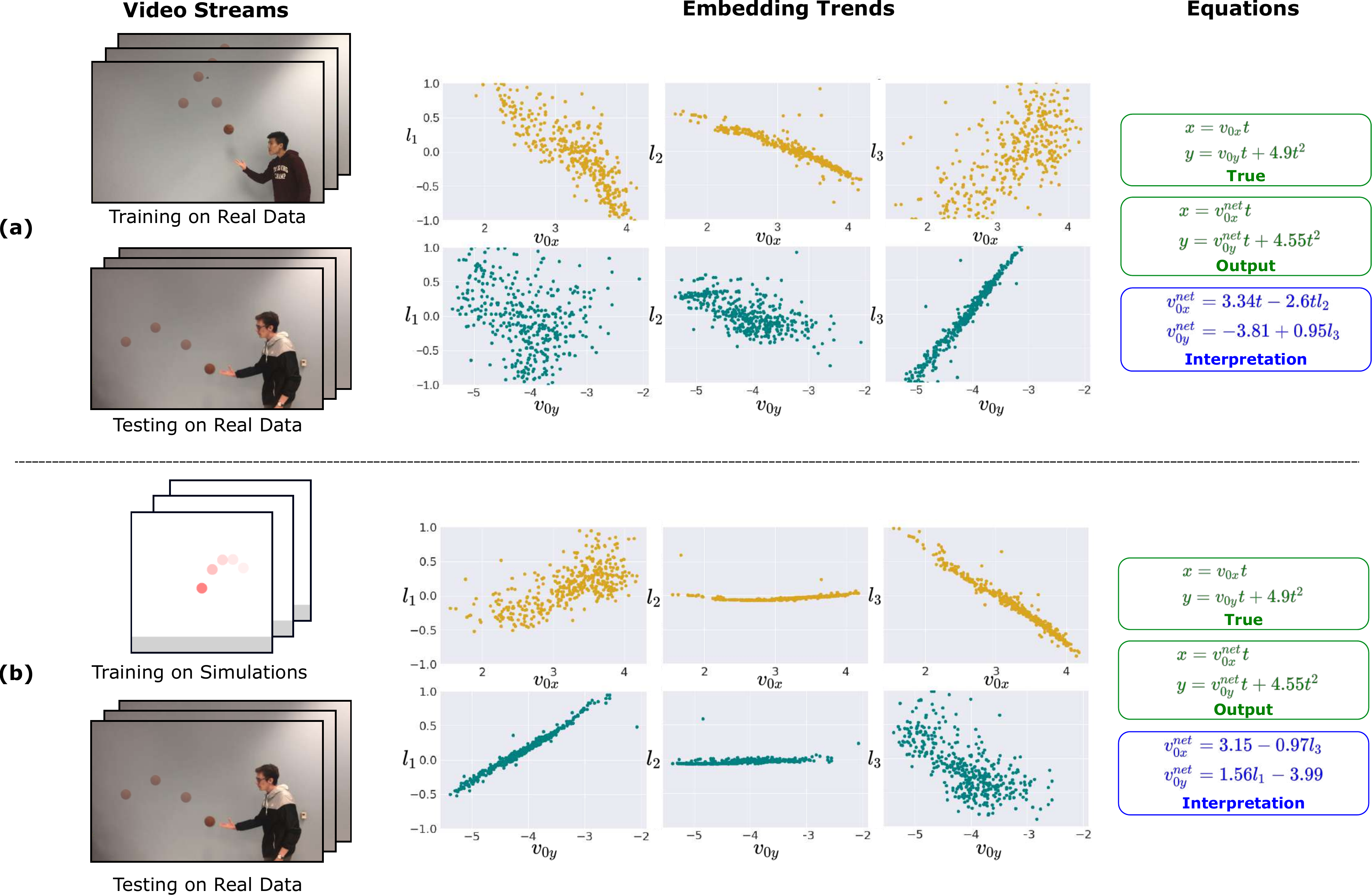}
    \caption{
    \textbf{Evaluating performance on real data, in two conditions.} (a) Testing on a set of real data, and training on real data. The videos of several basketball tosses are used as input to the pipeline. The accurate representations and the derived human interpretable equations, governing the real world phenomenon, are shown to emphasize the robustness of the pipeline. In (b), similar approach but the training set is synthetic data. Similar performance is observed, which underscores that the proposed results are not obtained from overfitting.}
    
    \label{fig:real_results}
\end{figure*}

Figure~\ref{fig:main_result} illustrates various results from our framework, tested on synthetically generated data described in Table~\ref{tab:dataset}. With \textsc{free fall}, we assess the ability of our system to perform with parameters that affect the discovery linearly (as coefficients to a term linear in time). With \textsc{constant acceleration}, we observe the performance on non-linear (quadratic) parameter effect. Finally, \textsc{circular motion} provides insight into performance for sinusoidal dependence. Results for two additional tasks, \textsc{helical motion} and \textsc{damped oscillation}, may be found in Appendix~\ref{app:hard_physics}.

\paragraph{\textsc{Free Fall} (synthetic):} In this scene, all possible trajectories are completely parameterized by the initial velocities $v_{0x}$ and $v_{0y}$ along the $x$ and $y$ directions. Figure~\ref{fig:main_result}(a) displays the output of our method for \textsc{free fall}, including both embeddings as well as the discovered equation. The embedding trends show that our latent physics model successfully learns to separate these horizontal and vertical velocity in two separate nodes. The correlation of the three latent nodes with the two governing (ground-truth) parameters demonstrate that the nodes learn an affine transform of the ground-truth velocities. It is important to note that the third node does not show dependence on the input, assuming a constant value. This reconciles with human intuition in the sense that \textsc{free fall} is determined only by two parameters. In evaluating the final output, we observe that the discovered governing equation matches the form of the familiar kinematic equations. The value of the acceleration due to gravity is learnt exactly and the parametric dependence of the equation on the initial velocities is accurate up to an affine transform. 

\paragraph{\textsc{Constant Acceleration Motion} (synthetic):} In this task, the trajectory is governed by a single parameter: the acceleration $a$ acting on the object. Obtained results are displayed in Figure~\ref{fig:main_result}(b). As we expect, since only one of the nodes is required to describe the phenomenon, the embedding trends show that two nodes are invariant to the input and learn an almost constant, low magnitude value. The other node, which is correlated to the input, learns acceleration. Turning to the output equations, we find our method discovers both the correct form, and the latent variable maps to an interpretation of $a$. Also note that the value of the $y$ coordinate, which is expected to be constant, is discovered accurately.

\paragraph{\textsc{Uniform Circular Motion} (synthetic):} This task has a sinusoidal, rather than polynomial form. For a fixed radius of revolution, the governing parameter we seek to discover is the angular frequency $\omega$ of the rotating object. Hence, this task also depends on a single governing parameter. Figure~\ref{fig:main_result}(c) highlights that one of the latent parameters is correlated with angular frequency, while the other two are uncorrelated to the input. Based on the learned parameters and observed positions, the proposed method correctly identifies a sinusoidal dependence for both the $x$ and the $y$ coordinates. 

\begin{figure*}
    \centering
    \includegraphics[width=\linewidth]{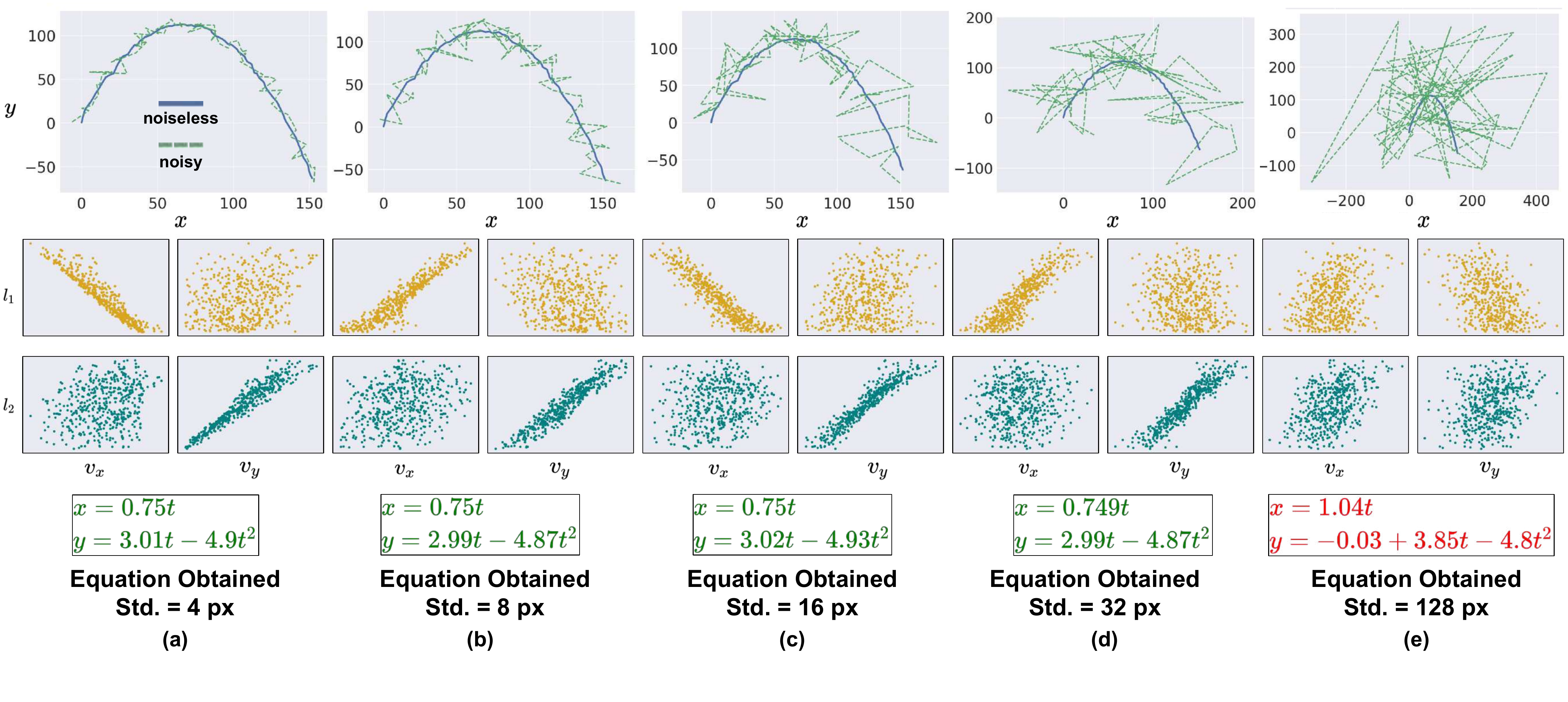}
    \caption{
    \textbf{The proposed method is found to be robust when considerable zero-mean additive Gaussian noise is added to the trajectory.} The pipeline is tested on synthetically added noise with standard deviation ranging from 4 to 128 pixels (at a scale of 300 pixels/meter). The representations are found to be robust for up to noise of standard deviation up to 32 pixels, with equations demonstrating analogous robustness. The method fails at a noise of standard deviation 128 pixels, which can be seen to completely bury the trajectory signal in noise.
    }
    \label{fig:noise_analysis}
\end{figure*}

\subsection{Real Data Evaluation} \label{ss:real} 

\paragraph{\textsc{free fall} (real experiment):} We replicate \textsc{free fall} in the real-world in a relatively uncontrolled manner. As shown in Figure~\ref{fig:real_results} the test set is a video sequence of a human tossing a ball with varying spins and uncontrolled air resistance. The motion may also not be perpendicular to the camera, leading to scale inconsistencies. 411 videos are collected, where each video represents a toss. To obtain ground truth initial velocities, we fit the kinematic equations to the observed videos, using the appropriate scaled value of the acceleration due to gravity $g$. The proposed latent discovery module does not have the luxury of this information. We report results in two conditions. In Figure~\ref{fig:real_results}(a), we train on real data and test on real data. Diversity in the dataset occurs due to different types of spins and tosses. To show that our method is not overfitting, Figure~\ref{fig:real_results}(b) displays results when we train on synthetic data and test on real data. Both cases achieve successful discovery of the ground-truth governing equation. In particular, two latent nodes show strong affine correlations with the ground truth horizontal and vertical velocities. In contrast, the third node, as we would expect, is uncorrelated (since only two parameters, $v_{0x}, v_{0y}$ govern the system). The symbolic form of the equation we learn reconciles with the known physics model up to an affine transform in the governing parameters. It is important to note that slight error is observed when testing on real data. In both Figure~\ref{fig:real_results}(a) and~\ref{fig:real_results}(b) the value of acceleration we learn due to gravity is off by a factor of about $7\%$. We believe the following reasons account for a part of this inconsistency: (i) Noise due to the greater Mask R-CNN error on the real videos, as compared to the simulations; and (ii) physical non-idealities such as air resistance and drag. We successfully test our method on an additional real task, \textsc{uniform circular motion}. Please refer to Appendix~\ref{app:real_scenes} for details and results.

\subsection{Performance Analysis} \label{ss:performance}

We now look at analyzing, in reasonable detail, the characteristics and performance of the proposed approach. These factors hold special importance towards the function of the pipeline as a physics discovery unit, in a future application domain (e.g. biomedical, astrophysics).

\paragraph{Latent nodes an affine transform of ground truth:} Figure~\ref{fig:main_result} and Figure~\ref{fig:real_results} explicitly show that the latent nodes are an affine transformation of the ground truth, governing parameters. This reinforces our claim that the latent parameters we learn are human interpretable. Due to the use of a $\beta$-VAE, the latent physics module is constrained to learn sparse representations, subject to a Pareto fit. Adding additional latent nodes therefore results in representations for these superfluous nodes either being entirely uncorrelated to the governing parameters, or of extremely low magnitude. The affine transform is important, not only for interpretability, but also because a linear least squares can be used to tune the parameters once the governing equation has been identified. 

\paragraph{Robustness against noise:} To assess performance in context of noise, we use the synthetic \textsc{free fall} task and add noise to the position detection module of varying strengths. This corrupted data is then used to train the latent physics module and serve as the input to the equation discovery module. The plots of governing parameters in Figure~\ref{fig:noise_analysis} show that with increasingly noisy input trajectories, the representations remain relatively robust. However, the variance in representations is found to increase as the input corruption level increases. We are satisfied with the quality of these representations. Using even noisy (yet correlated) representations in the equation discovery step, still enables us to recover output equations that are symbolically accurate. The method eventually fails for corruption with noise of standard deviation of 128. At this very high noise level, even the direction of the trajectory is changing (i.e. the ball appears to travel backward). We can observe this in the last column of Figure~\ref{fig:noise_analysis}.

\begin{figure}
    \centering
    \includegraphics[width=\linewidth]{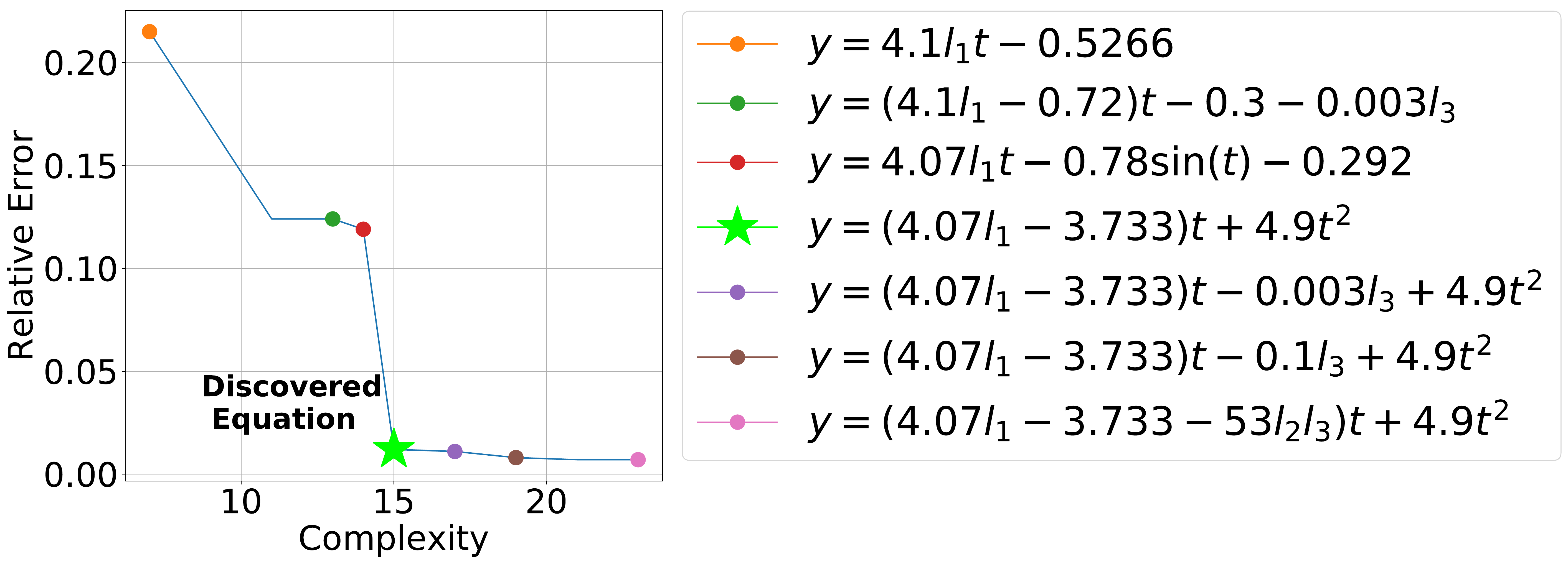}
    \caption{\textbf{Trade-off between equation complexity and accuracy.} We show multiple candidate equations for the synthetic free fall task along the vertical direction. The equation with the correct parametric form occurs at the optimal trade-off point.}
    \label{fig:trade-off}
\end{figure}

\paragraph{Equation complexity versus accuracy:} Here we discuss how the proposed framework is able to recover the correct equation by balancing optimality in context of \emph{equation sparsity} and \emph{performance fit}. The equation discovery module results in a set of possible equations, of varying complexity (a function of the number of terms and operations in the equation). In order to choose an appropriate trade-off between fitting accuracy and complexity, we use plots such as those shown in Figure~\ref{fig:trade-off}. The knee point of the trade-off curve is chosen as the expression of interest, since it marks the point of maximum gain in error performance with minimal increase in complexity. Such a selection ensures that the genetic programming algorithm refrains from over-fitting on the relevant data, which is essential towards allowing for interpretability. This is also analogous to similar observations from representation learning, where there is an understood trade-off between the extent of disentanglement of latent embeddings and downstream prediction accuracy~\cite{higgins2017betaVAE}.

\begin{figure}
    \centering
    \includegraphics[width=\linewidth]{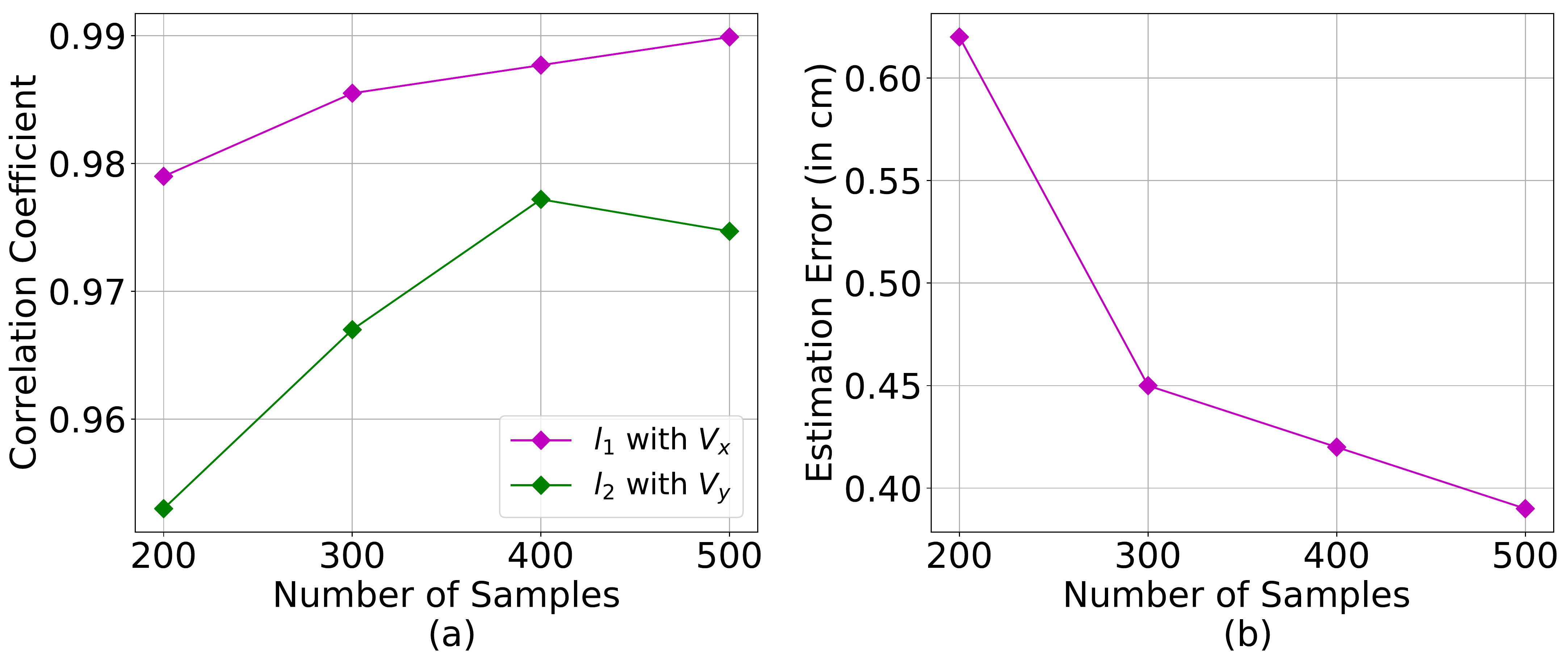}
    \caption{\textbf{Visual Physics framework improves consistently with different numbers of training samples.} We test the performance on the free-fall task under dataset sizes of 200, 300, 400, and 500 respectively. (a) shows the correlation coefficients between the ground-truth physical parameters and the discovered physical parameters, and (b) shows the mean squared error of the estimated locations in centimeters.}
    \label{fig:training_samples_effect}
\end{figure}

\paragraph{Effect of training data size:} Finally, we analyze the performance of our proposed method with respect to varying amounts of training data. This holds relevance in terms of the possible application of the pipeline (or others inspired by it) toward tasks with varying data availability. Figure~\ref{fig:training_samples_effect} shows the results of this analysis on the synthetic \textsc{free fall} task. We evaluate performance based on: (a) the normalized cross-correlation coefficient between the learnt active latent node and the ground-truth governing parameters, and (b) the trajectory prediction accuracy based on the latent values predicted by the latent physics module on test data, used on the discovered equations. Please refer to Appendix~\ref{app:second_order_implementation} for a detailed description of these metrics. The general trend of increasing correlation and reducing prediction error with increasing training samples is clearly visible in the plots. However, what is also of interest is the fact that the worst case error for the scenario with the lowest number of input samples (200 samples) has a sufficiently high correlation of 0.95. This highlights the versatility and robustness of the proposed approach towards a range of possible tasks. 

\section{Discussion}

In summary, we have demonstrated the ability to discover physics from video streams. Our method is unique in that it is able to discover both the governing equations and physical parameters. Our results are powered by an encoder-decoder framework that learns latent representations. We show that, even in cases of significant noise, the latent representations are physically interpretable.

\paragraph{Beyond 2D phenomena:} The Visual Physics dataset consists of 2-dimensional scenarios. For example, the tossing ball is viewed from the side, such that the ball does not change in its axial depth. For engineering reasons, we assume that the physical phenomena is observed in the 2D camera space of a video camera. If dynamics occur in 3-dimensions (e.g. motion in $x,y,z$), then our algorithmic pipeline is still valid, but we must use a 3D camera to capture these 3D dynamics. In general, Visual Physics framework can apply to higher-dimensional scenarios, potentially outside of video, provided that the measurement space is able to capture the phenomena. 

\paragraph{Applications:} For reader accessibility and experimental reproducibility, we have chosen simple problems (like projectile motion and circular motion). However, we could envision future applications of this framework to domains like high-energy astrophysics, optical scattering, and medical imaging where the governing equations are unknown or partially known. In medical imaging, for example, it is important to find latent embeddings that are both discriminative, but also physically interpretable.

\paragraph{Open problems:} Analogous to the apocryphal story of Newton's apple we have considered dynamics of a single object. This work is therefore a stepping stone to understanding the dynamics of multiple-objects. Another open problem is to extend the pipeline, beyond the three modules we have proposed. Concretely, we could also see adding a fourth module where the equation and embeddings we discover is used as input to another inference framework. For example, it might be possible to improve object detection given the velocities of objects, or create computational imaging pipelines that learn to classify scenes based on scattering properties. In conclusion, this paper is scratching the surface of the possibilities at the seamline of computer vision, physics, and artificial intelligence. We are excited to see these fields continue to merge.

{\small
\bibliographystyle{ieee_fullname}
\bibliography{egbib}

\begin{thebibliography}{10}\itemsep=-1pt

\bibitem{ba2019physics}
Yunhao Ba, Rui Chen, Yiqin Wang, Lei Yan, Boxin Shi, and Achuta Kadambi.
\newblock Physics-based neural networks for shape from polarization.
\newblock {\em arXiv preprint arXiv:1903.10210}, 2019.

\bibitem{Battalgia2016IN}
Peter Battaglia, Razvan Pascanu, Matthew Lai, Danilo Jimenez~Rezende, and Koray
  Kavukcuoglu.
\newblock Interaction networks for learning about objects, relations and
  physics.
\newblock {\em NeurIPS}, 2016.

\bibitem{Bhat2002}
Kiran~S. Bhat, Steven~M. Seitz, and Jovan Popovic.
\newblock Computing the physical parameters of rigid-body motion from video.
\newblock {\em ECCV}, 2002.

\bibitem{Brubaker2009}
Marcus~A. Brubaker, Leonid Sigal, and David~J. Fleet.
\newblock Estimating contact dynamics.
\newblock {\em ICCV}, 2009.

\bibitem{chen2018reblur2deblur}
Huaijin Chen, Jinwei Gu, Orazio Gallo, Ming-Yu Liu, Ashok Veeraraghavan, and
  Jan Kautz.
\newblock Reblur2deblur: Deblurring videos via self-supervised learning.
\newblock {\em ICCP}, 2018.

\bibitem{davis2015visual}
Abe Davis, Katherine~L Bouman, Justin~G Chen, Michael Rubinstein, Fredo Durand,
  and William~T Freeman.
\newblock Visual vibrometry: Estimating material properties from small motion
  in video.
\newblock {\em CVPR}, 2015.

\bibitem{Kaizen}
Vin\'{\i}cius~Veloso De~Melo.
\newblock Kaizen programming.
\newblock {\em Annual Conference on Genetic and Evolutionary Computation},
  2014.

\bibitem{diamond2017unrolled}
Steven Diamond, Vincent Sitzmann, Felix Heide, and Gordon Wetzstein.
\newblock Unrolled optimization with deep priors.
\newblock {\em arXiv preprint arXiv:1705.08041}, 2017.

\bibitem{Eslami1204GenerativeQueryNet}
S.~M.~Ali Eslami, Danilo Jimenez~Rezende, Frederic Besse, Fabio Viola, Ari~S.
  Morcos, Marta Garnelo, Avraham Ruderman, Andrei~A. Rusu, Ivo Danihelka, Karol
  Gregor, David~P. Reichert, Lars Buesing, Theophane Weber, Oriol Vinyals, Dan
  Rosenbaum, Neil Rabinowitz, Helen King, Chloe Hillier, Matt Botvinick, Daan
  Wierstra, Koray Kavukcuoglu, and Demis Hassabis.
\newblock Neural scene representation and rendering.
\newblock {\em Science}, 2018.

\bibitem{fei2019geo}
Xiaohan Fei, Alex Wong, and Stefano Soatto.
\newblock Geo-supervised visual depth prediction.
\newblock {\em IEEE Robotics and Automation Letters}, 2019.

\bibitem{felzenszwalb2008discriminatively}
Pedro Felzenszwalb, David McAllester, and Deva Ramanan.
\newblock A discriminatively trained, multiscale, deformable part model.
\newblock {\em ICCV}, 2008.

\bibitem{felzenszwalb2009object}
Pedro~F Felzenszwalb, Ross~B Girshick, David McAllester, and Deva Ramanan.
\newblock Object detection with discriminatively trained part-based models.
\newblock {\em PAMI}, 2009.

\bibitem{Fragidaki16Billiards}
Katerina Fragkiadaki, Pulkit Agrawal, Sergey Levine, and Jitendra Malik.
\newblock Learning visual predictive models of physics for playing billiards.
\newblock {\em ICLR}, 2016.

\bibitem{gregor2010learning}
Karol Gregor and Yann LeCun.
\newblock Learning fast approximations of sparse coding.
\newblock {\em ICML}, 2010.

\bibitem{Halder_2019_ICCV}
Shirsendu~Sukanta Halder, Jean-Francois Lalonde, and Raoul~de Charette.
\newblock Physics-based rendering for improving robustness to rain.
\newblock {\em ICCV}, 2019.

\bibitem{he2017mask}
Kaiming He, Georgia Gkioxari, Piotr Doll{\'a}r, and Ross Girshick.
\newblock Mask {R-CNN}.
\newblock {\em ICCV}, 2017.

\bibitem{higgins2017betaVAE}
Irina Higgins, Loic Matthey, Arka Pal, Christopher Burgess, Xavier Glorot,
  Matthew Botvinick, Shakir Mohamed, and Alexander Lerchner.
\newblock $\beta$-vae: Learning basic visual concepts with a constrained
  variational framework.
\newblock {\em ICLR}, 2017.

\bibitem{hills2015algorithm}
Daniel~JA Hills, Adrian~M Gr{\"u}tter, and Jonathan~J Hudson.
\newblock An algorithm for discovering lagrangians automatically from data.
\newblock {\em PeerJ Computer Science}, 2015.

\bibitem{horn1989shape}
Berthold~KP Horn and Michael~J Brooks.
\newblock {\em Shape from shading}.
\newblock 1989.

\bibitem{huang2018NIPSworkshop}
Siyu Huang, Zhi-Qi Cheng, Xi Li, Xiao Wu, Zhongfei Zhang, and Alexander
  Hauptmann.
\newblock Perceiving physical equation by observing visual scenarios.
\newblock {\em arXiv preprint arXiv:1811.12238}, 2018.

\bibitem{ikeuchi1981numerical}
Katsushi Ikeuchi and Berthold~KP Horn.
\newblock Numerical shape from shading and occluding boundaries.
\newblock {\em Artificial Intelligence}, 1981.

\bibitem{iten2018discovering}
Raban Iten, Tony Metger, Henrik Wilming, L{\'\i}dia Del~Rio, and Renato Renner.
\newblock Discovering physical concepts with neural networks.
\newblock {\em arXiv preprint arXiv:1807.10300}, 2018.

\bibitem{jin2017deep}
Kyong~Hwan Jin, Michael~T McCann, Emmanuel Froustey, and Michael Unser.
\newblock Deep convolutional neural network for inverse problems in imaging.
\newblock {\em IEEE Transactions on Image Processing}, 2017.

\bibitem{kang2017deep}
Eunhee Kang, Junhong Min, and Jong~Chul Ye.
\newblock A deep convolutional neural network using directional wavelets for
  low-dose x-ray ct reconstruction.
\newblock {\em Medical Physics}, 2017.

\bibitem{kass1988snakes}
Michael Kass, Andrew Witkin, and Demetri Terzopoulos.
\newblock Snakes: Active contour models.
\newblock {\em IJCV}, 1988.

\bibitem{kingma2014adam}
Diederik~P Kingma and Jimmy Ba.
\newblock Adam: A method for stochastic optimization.
\newblock {\em arXiv preprint arXiv:1412.6980}, 2014.

\bibitem{kingma2013auto}
Diederik~P Kingma and Max Welling.
\newblock Auto-encoding variational bayes.
\newblock {\em arXiv preprint arXiv:1312.6114}, 2013.

\bibitem{GeneticProgramming}
Koza and John R.
\newblock Genetic programming: On the programming of computers by means of
  natural selection.
\newblock {\em MIT Press}, 1992.

\bibitem{NeuralSymbolicRegression2019}
Li {Li}, Minjie {Fan}, Rishabh {Singh}, and Patrick {Riley}.
\newblock {Neural-Guided Symbolic Regression with Semantic Prior}.
\newblock {\em arXiv preprint arXiv:1901.07714}, {2019}.

\bibitem{li2019restoration}
Ruoteng Li, Loong-Fah Cheong, and Robby~T. Tan.
\newblock Heavy rain image restoration: Integrating physics model and
  conditional adversarial learning.
\newblock {\em CVPR}, 2019.

\bibitem{lin2014microsoft}
Tsung-Yi Lin, Michael Maire, Serge Belongie, James Hays, Pietro Perona, Deva
  Ramanan, Piotr Doll{\'a}r, and C~Lawrence Zitnick.
\newblock Microsoft coco: Common objects in context.
\newblock {\em ECCV}, 2014.

\bibitem{EQL}
Georg Martius and Christoph~H. Lampert.
\newblock Extrapolation and learning equations.
\newblock {\em arXiv preprint arXiv:1610.02995}, 2016.

\bibitem{maslyaev2019data}
Michail Maslyaev, Alexander Hvatov, and Anna Kalyuzhnaya.
\newblock Data-driven pde discovery with evolutionary approach.
\newblock {\em arXiv preprint arXiv:1903.08011}, 2019.

\bibitem{Mottaghi15Newton}
Roozbeh Mottaghi, Hessam Bagherinezhad, Mohammad Rastegari, and Ali Farhadi.
\newblock Newtonian image understanding: Unfolding the dynamics of objects in
  static images.
\newblock {\em CoRR}, abs/1510.04048, 2015.

\bibitem{o2014computational}
Matthew O'Toole and Gordon Wetzstein.
\newblock Computational cameras and displays.
\newblock {\em ACM SIGGRAPH}, 2014.

\bibitem{o2018confocal}
Matthew O’Toole, David~B Lindell, and Gordon Wetzstein.
\newblock Confocal non-line-of-sight imaging based on the light-cone transform.
\newblock {\em Nature}, 2018.

\bibitem{paszke2017automatic}
Adam Paszke, Sam Gross, Soumith Chintala, Gregory Chanan, Edward Yang, Zachary
  DeVito, Zeming Lin, Alban Desmaison, Luca Antiga, and Adam Lerer.
\newblock Automatic differentiation in {PyTorch}.
\newblock {\em NeurIPS Autodiff Workshop}, 2017.

\bibitem{purushwalkam2019bounce}
Senthil Purushwalkam, Abhinav Gupta, Danny~M Kaufman, and Bryan Russell.
\newblock Bounce and learn: Modeling scene dynamics with real-world bounces.
\newblock {\em arXiv preprint arXiv:1904.06827}, 2019.

\bibitem{GP-RVM}
Hossein~Izadi Rad, Ji Feng, and Hitoshi Iba.
\newblock {GP-RVM:} genetic programing-based symbolic regression using
  relevance vector machine.
\newblock {\em CoRR}, abs/1806.02502, 2018.

\bibitem{ramesh20085d}
Raskar Ramesh and James Davis.
\newblock 5d time-light transport matrix: What can we reason about scene
  properties?
\newblock Technical report, 2008.

\bibitem{Brunton2016}
Samuel~H. Rudy, Steven~L. Brunton, Joshua~L. Proctor, and J.~Nathan Kutz.
\newblock Data-driven discovery of partial differential equations.
\newblock {\em Science}, 2016.

\bibitem{EQL_extented}
Subham~S. Sahoo, Christoph~H. Lampert, and Georg Martius.
\newblock Learning equations for extrapolation and control.
\newblock {\em ICML}, 2018.

\bibitem{EureqaSoftware}
Lipson~H. Schmidt, M.
\newblock Eureqa (version 0.98 beta) [software].
\newblock 2014.

\bibitem{schmidt2009distilling}
Michael Schmidt and Hod Lipson.
\newblock Distilling free-form natural laws from experimental data.
\newblock {\em Science}, 2009.

\bibitem{shi2019neural}
Guanya Shi, Xichen Shi, Michael O’Connell, Rose Yu, Kamyar Azizzadenesheli,
  Animashree Anandkumar, Yisong Yue, and Soon-Jo Chung.
\newblock Neural lander: Stable drone landing control using learned dynamics.
\newblock {\em ICRA}, 2019.

\bibitem{stewart2017label}
Russell Stewart and Stefano Ermon.
\newblock Label-free supervision of neural networks with physics and domain
  knowledge.
\newblock {\em AAAI}, 2017.

\bibitem{tanaka2018material}
Kenichiro Tanaka, Yasuhiro Mukaigawa, Takuya Funatomi, Hiroyuki Kubo, Yasuyuki
  Matsushita, and Yasushi Yagi.
\newblock Material classification from time-of-flight distortions.
\newblock {\em PAMI}, 2018.

\bibitem{velten2012recovering}
Andreas Velten, Thomas Willwacher, Otkrist Gupta, Ashok Veeraraghavan, Moungi~G
  Bawendi, and Ramesh Raskar.
\newblock Recovering three-dimensional shape around a corner using ultrafast
  time-of-flight imaging.
\newblock {\em Nature Communications}, 2012.

\bibitem{wang2019deriving}
Youchao Wang, Sam Willis, Vasileios Tsoutsouras, and Phillip Stanley-Marbell.
\newblock Deriving equations from sensor data using dimensional function
  synthesis.
\newblock {\em ACM Transactions on Embedded Computing Systems (TECS)}, 2019.

\bibitem{Watters2017VisualInteractionNetworks}
Nicholas Watters, Daniel Zoran, Theophane Weber, Peter Battaglia, Razvan
  Pascanu, and Andrea Tacchetti.
\newblock Visual interaction networks: Learning a physics simulator from video.
\newblock {\em NeurIPS}, 2017.

\bibitem{winkler2005new}
Stephan Winkler, Michael Affenzeller, and Stefan Wagner.
\newblock New methods for the identification of nonlinear model structures
  based upon genetic programming techniques.
\newblock {\em SYSTEMS SCIENCE-WROCLAW-}, 2005.

\bibitem{woodham1980photometric}
Robert~J Woodham.
\newblock Photometric method for determining surface orientation from multiple
  images.
\newblock {\em Optical Engineering}, 1980.

\bibitem{Wu2017Deanimation}
Jiajun Wu, Erika Lu, Pushmeet Kohli, Bill Freeman, and Josh Tenenbaum.
\newblock Learning to see physics via visual de-animation.
\newblock {\em NeurIPS}, 2017.

\bibitem{JiajunWu2015Gallileo}
Jiajun Wu, Ilker Yildirim, Joseph~J Lim, Bill Freeman, and Josh Tenenbaum.
\newblock Galileo: Perceiving physical object properties by integrating a
  physics engine with deep learning.
\newblock {\em NeurIPS}, 2015.

\bibitem{xin2019theory}
Shumian Xin, Sotiris Nousias, Kiriakos~N Kutulakos, Aswin~C Sankaranarayanan,
  Srinivasa~G Narasimhan, and Ioannis Gkioulekas.
\newblock A theory of fermat paths for non-line-of-sight shape reconstruction.
\newblock {\em CVPR}, 2019.

\bibitem{zeng2019tossingbot}
Andy Zeng, Shuran Song, Johnny Lee, Alberto Rodriguez, and Thomas Funkhouser.
\newblock Tossingbot: Learning to throw arbitrary objects with residual
  physics.
\newblock {\em arXiv preprint arXiv:1903.11239}, 2019.

\end{thebibliography}
}

\newpage
\onecolumn
\appendix

\begin{center}
    \LARGE\textbf{Supplementary Results}
    \vspace{0.5cm}
\end{center}

This supplement is organized as follows:
\begin{enumerate}[itemsep=0em]
    \item Appendix~\ref{app:real_scenes} includes a real scene with a sinusoidal, rather than polynomial, physical form.
    \item Appendix~\ref{app:hard_physics} shows that the method generalizes to more difficult physical problems, in context of mathematical form (e.g. exponential decay, helical motion).  
    \item Appendix~\ref{app:second_order_implementation} discusses the quantitative metrics for performance evaluation. 
    \item Appendix~\ref{app:software_implementation} describes specific implementation details and includes source code snippets for key portions of the paper.
\end{enumerate}

\section{Circular Motion (real experiment)} \label{app:real_scenes}
\begin{figure}[h]
    \centering
    \includegraphics[width=\textwidth]{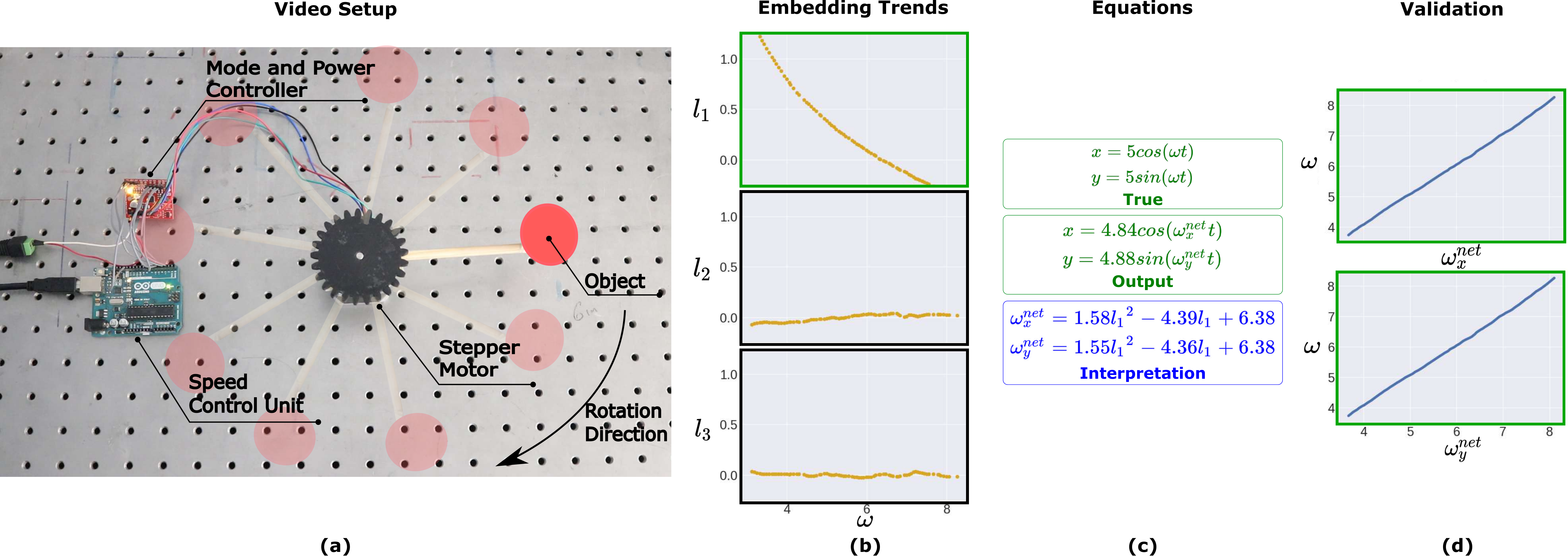}
    
    \caption{\textbf{Performance of Visual Physics framework on real circular motion.} The governing parameter is appropriately obtained and an interpretable governing equation reconciling with known equations is discovered. The interpretations from the discovered equations are validated to confirm their parameteric correspondence with $\omega$ in the ground-truth circular motion equations.}
    \label{fig:real_circular_results}
\end{figure}

Having discovered the equations for \textsc{circular motion} from synthetic data, experiments on this task are now extended to real data. Through this, we aim to further demonstrate the applicability of our method on real scenes. The dataset consists of 80 videos of an object rotating at fixed angular velocity. The rotation radius is kept constant across the dataset, and the angular velocity $\omega$ is varied in the range [$1.2\pi$, $3\pi$]~radians/s. Videos with $\omega<1.2\pi$ are excluded from the dataset in order to avoid non-linear effects of the motor at low frequencies. The first 200 frames of every video are used as input to the position detection module. The positions obtained are corrected for initial phase, so that all input trajectories have the same (zero) phase, by appropriate rotation of coordinates. The ground-truth $\omega$ for each video is calculated numerically based on these detected locations, from zero-crossing frequencies. These are used for verification of the learned representations, and are not used as part of the discovery process. Figure~\ref{fig:real_circular_results}(a) shows a graphical description of the setup for data collection.

The latent physics module is trained with synthetic data, which is generated so as to match the parameters of the real dataset (frame rate, angular velocity range). We then use the real data on this trained model, in order to obtain the latent representations and the inputs for the equation discovery module. It may be observed from Figure~\ref{fig:real_circular_results}(b) that the first latent embedding $l_1$ obtained for the real data is well-correlated with $\omega$. The other two nodes are close to zero in magnitude. This reconciles with the fact that there exists only one primary governing parameter for this setup. Additionally, the trend between the learnt embedding $l_1$ and $\omega$ suggests a quadratic relation. Hence, in Figure~\ref{fig:real_circular_results}(d), we verify that the discovered angular velocity $\omega^{net}$ (mentioned in Figure~\ref{fig:real_circular_results}(c)) corresponds to ground-truth $\omega$ with high accuracy. 

Here, it is important to emphasize the correlation of latent nodes with ground truth parameters, as shown in Figure~\ref{fig:training_samples_effect}. The interpretability of the discovered equations is directly related to the value of this correlation coefficient. This is easily evident in the affine mapping obtained between the latent parameters and underlying physics concepts, for the results in the main paper. However, we impose no such explicit linearity constraint in the pipeline, since that may be construed as prior human knowledge. As long as the proposed method learns representations that are strongly correlated with the underlying physics parameters, the discovered equations will be interpretable and will embody the physics parameters. Therefore, even if the latent embeddings have a quadratic (or any non-linear one-to-one) relationship with the ground truth, as observed for the \textsc{uniform circular motion} task for real scenes, interpretability is still maintained.

\section{Helical Motion and Damped Oscillation (synthetic scenes)} \label{app:hard_physics}

\begin{figure}[h]
    \centering
    \includegraphics[width = \textwidth]{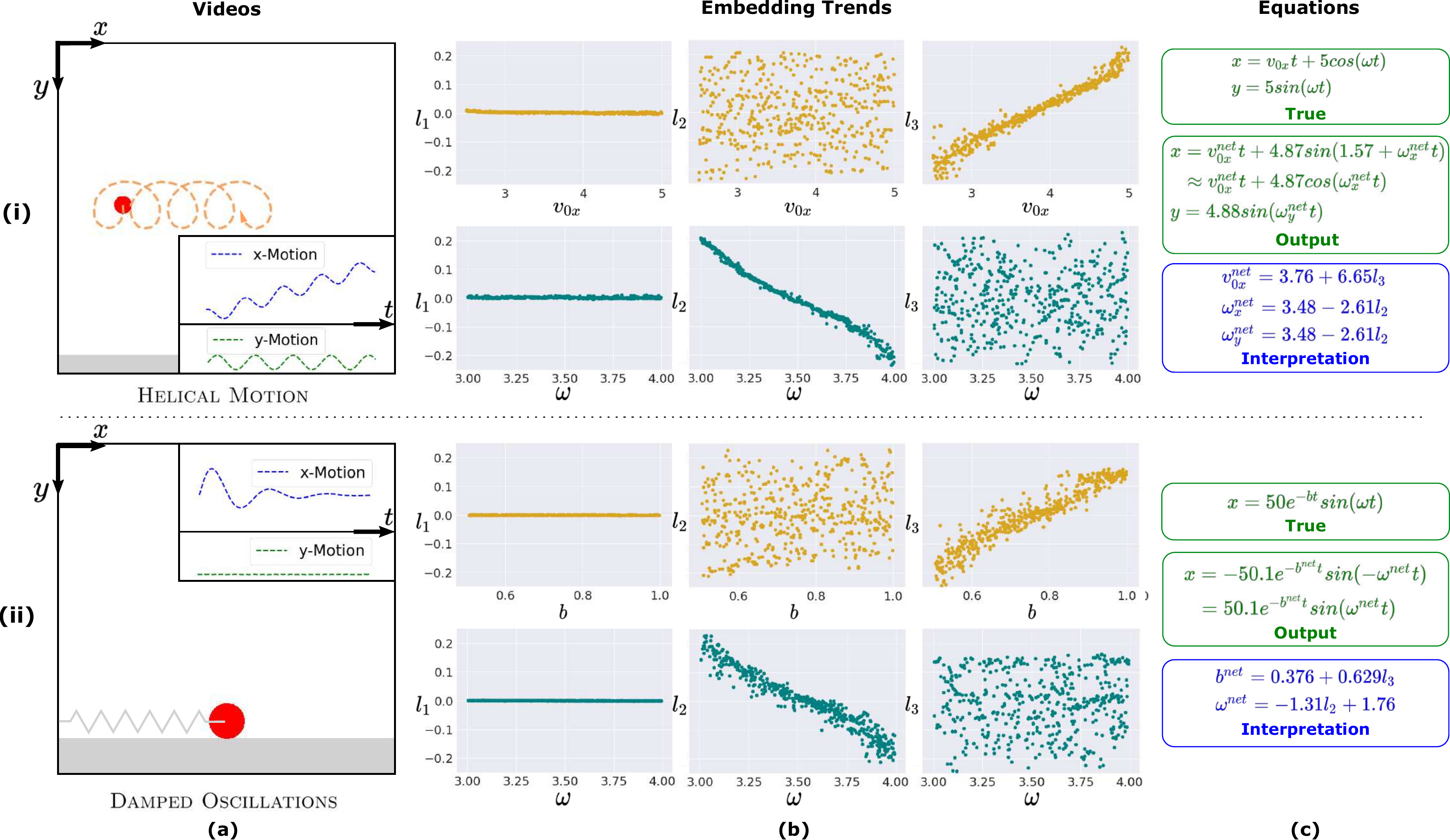}
    \caption{\textbf{Performance of Visual Physics framework on additional synthetic scenes.} Independent governing parameters can be discovered accurately from complex physical phenomena, and the corresponding equations are consistent with true physical expressions.}
    \label{fig:additional_syn_results}
\end{figure}

The evaluation on two additional synthetic tasks of greater difficulty in terms of functional form (\textsc{helical motion} and \textsc{damped oscillation}) is presented in this section. The corresponding results are illustrated in Figure~\ref{fig:additional_syn_results}.

\paragraph{\textsc{Helical Motion} (synthetic):} We demonstrate the discovery of translational and rotational motion in the main paper through the \textsc{free fall} and \textsc{uniform circular motion} datasets. To increase the complexity of the physics task, we now evaluate the proposed framework for 2-dimensional helical motion, where both translational and rotational motion act together. The synthetic videos are generated with different angular velocities $\omega$ and horizontal translational velocities $v_{0x}$. There is no translational motion along the $y$ direction, and the radius of the rotational motion is held constant for all the videos. Of the 600 videos in this synthetic dataset, 500 are used for training. Figure~\ref{fig:additional_syn_results}(i) shows the learnt representations and equations along the $x$ and $y$ directions. It may be observed that two of the latent representations are affine transforms of the governing physical parameters, $v_{0x}$ and $\omega$, and the derived equations are of the same functional form as the true equations. This emphasizes the performance of our framework on scenarios with multiple physical phenomena in action.

\paragraph{\textsc{Damped Oscillation} (synthetic):} Damping is a general energy loss mechanism for various systems, and one of the common forms of damping is the exponential decay. In this experiment, we simulate videos of damped oscillation, where the oscillation amplitude decays exponentially with time. We aim to test the capability of the proposed method towards discovering physical laws of more complex forms. We only change the damping factor $b$ and the angular frequency $\omega$ along $x$ direction, while the object location along $y$ direction is fixed. 600 videos are generated with random initial conditions as part of the dataset. Among these, 500 are used to train the proposed architecture, and the remaining constitute the test set. As shown in Figure~\ref{fig:additional_syn_results}(ii), the latent physics module is able to discover the notion of $\omega$ and $b$ in two different nodes, and the equation discovery module can generate equations to describe the combination of periodic and damped motions accurately. 

\section{Quantitative Performance Evaluation} \label{app:second_order_implementation}
The performance of the proposed Visual Physics framework may be measured along two fronts: (i) the mean error between the ground-truth trajectories and the trajectories from discovered equations, and (ii) the normalized cross-correlation coefficient between the latent representations and the corresponding ground-truth governing parameters. The analysis on the effect of training data size, from onward in the main paper, utilizes these metrics for evaluation. Here, we describe these metrics in more detail.

Let the ground-truth trajectory coordinates be denoted by $(x^{(t)}, y^{(t)})$ at a given time instant $t$. Based on the Visual Physics framework, let the learnt equations for $x$ and $y$ be given by $x = f_x(t, l_1, l_2, ..., l_n)$ and $y = f_y(t, l_1, l_2, ..., l_n)$, where $l_1, l_2, ..., l_n$ are the latent node values. Then, the mean error between trajectories ($\epsilon$) can be computed as 
\begin{equation}
    \epsilon=\sqrt{\frac{\Sigma_t(x^{(t)}-f_x(t,l_1,l_2,...,l_n))^2}{S}+\frac{\Sigma_t(y^{(t)}-f_y(t,l_1,l_2,...,l_n))^2}{S}},
\end{equation}
where $S$ is the total number of time samples in the trajectory under consideration. Additionally, the values for $l_1, l_2, ..., l_n$ are estimated through least-squares. Some values of $\epsilon$ evaluated on trajectories for the \textsc{free fall} case may be found in Figure~\ref{fig:training_samples_effect}. A test set of unseen trajectories was evaluated using these metrics. A low value of the error implies that the model (equation) learnt is sufficiently parametrized to characterize the observed trajectory, as well as that the time evolution of the predicted trajectory matches that of the observed trajectory.

Let the ground-truth governing parameters be represented by $g_1, g_2, ..., g_m$, $m \leq n$. On successful discovery, the hidden nodes of the latent physics module are expected to show strong correlations with the governing parameters. Hence, the normalized cross-correlation between corresponding latent nodes and governing parameters is given by 
\begin{equation}
    C_{i,j}=\frac{\Sigma_{k=1}^K g_i^{(k)}l_j^{(k)}}{K\sigma_{g_i}\sigma_{l_j}},
\end{equation}
where $K$ is the number of test trajectories, and $\sigma_u$ is the standard deviation of the variable $u$. We look at the magnitude of the strongly correlated hidden node-governing parameter pairs, and use the magnitude as an indicator of `goodness of latent representations'. Figure~\ref{fig:training_samples_effect} again highlights the computed values of the same for the \textsc{free fall} task. It may be observed that the values of the correlation metric are acceptably high. An additional metric for the goodness of latent representations and complexity evaluation can be the number of latent nodes required for the task. For instance, it would be interesting to apply this framework on multi-dimensional physics tasks, where the governing parameters are a lot more than 3, requiring us to use more number of latent parameters. 

\section{Software Implementation Details} \label{app:software_implementation}
This section highlights the synthetic dataset generation and pipeline implementation. We provide reproducible code snippets for one of the synthetic tasks, \textsc{free fall}. 

\paragraph{Dataset Generation (synthetic data):} The synthetic dataset comprises of an object undergoing motions governed by a range of diverse physical laws. We use Python and associated toolkits for simulating the same phenomena. Specifically, we use NumPy (np) and OpenCV (cv2). Each scene consists of a spherical object, of fixed size. The code for generating the object is shown below.
\begin{python}
    def draw_object(radius=25, color):
        ball = np.zeros((2*radius+1,2*radius+1,3))
        ball = cv2.circle(ball, (int(radius),int(radius)), radius, color, -1) 
        ball = ball.astype(np.float32)/255
        return ball
\end{python}

The background is chosen to be a constant frame, independent of the video. Frame rate, video duration and frame size are the tunable parameters for this setup. The trajectory of the ball is then calculated, based on initial positions, initial velocities and time. Specifically, the initial velocity range is chosen so that for a given initial position, the object always stays in the frame at all times. The code snippet for the same is as follows. 

\begin{python}
    yRng = [np.floor(3*frameSize[0]/5),frameSize[0]-halfSize-1-factor]
    xRng = [halfSize+1+factor,np.floor(frameSize[1]/5)]
    initPt = np.array([np.random.uniform(yRng[0],yRng[1]),np.random.uniform(xRng[0],xRng[1])])
    xVel = ((frameSize[1]-initPt[1]-halfSize)/Tgen)
    yVelUp = np.sqrt((initPt[0]-halfSize)*2*g)-1
    yVelDown = (((0.5*g*Tgen*Tgen)-(frameSize[0]-initPt[0]-halfSize))/Tgen)+1
    V_I = np.array([-1*np.random.uniform(yVelDown,yVelUp),np.random.uniform(0,xVel)])
\end{python}

Based on these parameters, the object location at each time instant is determined using the kinematic equations, and the corresponding frame is created. Code for the same is below.

\begin{python}
    x_pos = initPt[1]+(V_I[1]*time)
    y_pos = initPt[0]+(V_I[0]*time)+(0.5*g*time*time)
    frame = background
    temp = frame[int(y_pos-halfSize):int(y_pos+halfSize)+1,int(x_pos-halfSize):int(x_pos+halfSize)+1]
    frame[int(y_pos-halfSize):int(y_pos+halfSize)+1,int(x_pos-halfSize):int(x_pos+halfSize)+1]=(temp*maskBar)+(ball*mask) 
\end{python}

These sets of frames are then stored as the respective videos. Note that for the train on simulated, test on real regime for the \textsc{uniform circular motion} and \textsc{free fall} tasks, the frame size, frame rate and scale were chosen so as to be consistent with the real data. 

\paragraph{Position Detection Module:}
To process the videos, we developed a Mask R-CNN~\cite{he2017mask} based pipeline to convert the videos into position vectors which can be processed by the latent physics module. The input to the Mask R-CNN is a video with $N$ frames ($N = 200$ for synthetic data). The frames are sampled alternately, in a way that the even numbered frames are processed. The odd numbered frames are then used as the query input set. The output of the module is hence a $N + 1$ length vector, where the first $\frac{N}{2}$ elements correspond to $y$ coordinates, the next $\frac{N}{2}$ elements correspond to $x$ coordinates, and the last element of the vector corresponds to the frame number which is the query. The code snippet below illustrates the function which comprises the core of the position detection module. 
\begin{python}
    def rcnn_parameter_extractor(videoPath, numFrames, samp_rate, query_id, display=False, from_video=False, remove=0, subtract_firstframe=False):
        model = torchvision.models.detection.maskrcnn_resnet50_fpn(pretrained=True).cuda()
        model.eval(); centroid = []; flag = 0;
        label_dict = labels()
        mat = scipy.io.loadmat(videoPath)['vidTens']
        for i in range(remove, numFrames, samp_rate):
            image_tensor = torchvision.transforms.functional.to_tensor(mat[:,:,:,i]).cuda()
            output_image = model([image_tensor.float()])
            centroid.append(find_coord(output_image,mask=True))
            del image_tensor
        image_tensor = torchvision.transforms.functional.to_tensor(mat[:,:,:,query_id]).cuda()
        output_query = model([image_tensor.float()])
        centroid.append(find_coord(output_query,mask=True))
        cen = np.array(centroid)
        torch.cuda.empty_cache()
        del image_tensor
        return cen
\end{python}

For handling real data, the positions in the video were mapped from pixel coordinates to real world coordinates. In case of the \textsc{uniform circular motion} task, the position detection module was modified slightly to avoid unwanted detections by the Mask R-CNN in the video frames. The modification is to convolve each frame of the video with a Gaussian blur kernel (using OpenCV), so that other irrelevant stationary components of the video frame are partially  abstracted out, and the Mask R-CNN detects only the object of interest in the frame. Since we deal with a single object setting in our work, the blurring technique is useful to improve the robustness of the Mask R-CNN for detecting the object of interest in a variety of real scenes.

\paragraph{Latent Physics Module:} This module uses the position outputs from the previous step to identify governing parameters in the latent nodes. We use a feed-forward neural network for this purpose, specifically a modified $\beta$-Variational Auto Encoder ($\beta$-VAE) architecture~\cite{higgins2017betaVAE, Eslami1204GenerativeQueryNet}. The inputs of length $N+1$ are obtained from the position detection module ($N$ is the number of input video frames to the position detection module). Both the encoder and decoder consist of 6 fully-connected layers each. The dimension of all of the hidden layers is fixed at 256, and we use three latent nodes. For training, we concatenate a randomly chosen time query with the latent nodes, as input to the decoder. The outputs of the decoder are the position of the object at the specified time query. We use the mean squared error (MSE) loss on the predicted locations, regularized by the $\beta$-VAE disentanglement loss.

\paragraph{Equation Discovery Module:} We use the genetic programming toolkit Eureqa~\cite{schmidt2009distilling} to drive the equation discovery module. For our experiments, we use a fixed configuration setup. Inputs for the genetic programming are the positions along $x$ and $y$, the time instants $t$ (evaluated using the frame index and the frames rate of the video) and the latent node information for each trajectory $l_1,l_2,...,l_n$, where $n$ is the number of latent nodes used. For our experiments we use $n=3$. For $M$ training trajectories and $K$ samples per trajectory, we therefore have $M\times K$ sets of $(x, y, t, l_1, l_2, ..., l_n)$ as inputs. 

The error metric is chosen to be the R-squared goodness of fit. The candidate functions are chosen to be: (i) constant, (ii) input variable, (iii) addition, (iv) subtraction, (v) multiplication, (vi) division, (vii) sine, (viii) cosine and (ix) exponential. The complexity for each of the candidate functions is kept at the default value. No other configuration parameters for the toolkit are changed. Since the toolkit output includes several equations with varying complexities, the final equation is chosen based on pareto-optimality in the fit-complexity space.

\paragraph{Runtime Analysis:} 
Experiments were performed using a Linux (Ubuntu 18.04 LTS) machine with an Intel i5-8400 CPU (6 cores, 2.80 GHz), 16GB of RAM, and NVIDIA GeForce RTX 2070 GPU (8 GB of GPU RAM). Table~\ref{tab: runtime} shows the runtime analysis for the \textsc{helical motion} task. As suggested from the table, the overall runtime for this task is approximately 1.5 hours. The position detection module is the primary bottleneck in our pipeline, largely due to the size of the dataset. Depending on the complexity of the equation, the time required by the equation discovering module to converge at a plausible equation ranges from 60 s to 1800 s, for equations along two dimensions.  

\begin{table}[h]
\begin{center}
\begin{tabular}{lll}
\toprule
Module                                  & Runtime per unit          & Number of Units \\ \midrule
Position Detection Module               & 11 s per video            & 500 videos in a training set\\
Latent Physics Module                   & 60 s per 1000 epochs      & 2000 epochs required for convergence \\
Equation Discovery Module               & 30 s per equation         & 2 equations ($x$ and $y$ directions) \\ \midrule
\textbf{Overall Time}                   &                           & \textbf{5680 s} \\ \bottomrule
\end{tabular}
\end{center}
\caption{
\textbf{Runtime details of Visual Physics framework for \textsc{helical motion} task.}  The \textit{time} library in Python was used to compute the execution time for the position detection and latent physics module. The runtime for equation search was computed using the time stamps in the log-file of Eureqa.} 
\label{tab: runtime}
\end{table}

\end{document}